\documentclass{article} % For LaTeX2e
\usepackage{analemma,times}

\usepackage{amsmath,amsfonts,bm}

\def\eqref#1{equation~\ref{#1}}
\def\1{\bm{1}}

\DeclareMathAlphabet{\mathsfit}{\encodingdefault}{\sfdefault}{m}{sl}
\SetMathAlphabet{\mathsfit}{bold}{\encodingdefault}{\sfdefault}{bx}{n}

\usepackage[colorlinks=true,breaklinks=true]{hyperref}
\usepackage{url}
\usepackage{graphicx}
\usepackage{subfigure}
\usepackage{booktabs}
\usepackage{adjustbox}

\usepackage{amsmath}
\usepackage{amssymb}
\usepackage{mathtools}
\usepackage{amsthm}

\usepackage{multirow}
\usepackage{color}
\usepackage{colortbl}
\usepackage[capitalize,noabbrev]{cleveref}
\usepackage{xspace}

\definecolor{disclosurered}{rgb}{0.70,0.07,0.07}
\hypersetup{linkcolor=black,citecolor=black,urlcolor=black}

\theoremstyle{plain}

\theoremstyle{definition}

\theoremstyle{remark}

\graphicspath{{figures/}} % To reference your generated figures, name the PNGs directly.

\title{Position Bias Correction is Insufficient for One-Pass Attention Sorting}

\author{\textbf{FARS}, \\
\textbf{Qiong Tang}\footnotemark[2],
\textbf{Xiangkun Hu}\footnotemark[2],
\textbf{Xiangyang Liu}\footnotemark[2],
\textbf{Yiran Chen}\footnotemark[2],
\textbf{Yunfan Shao}\footnotemark[2] \\
Analemma \\
\texttt{fars@analemma.ai}
}

\analemmafinalcopy % Camera-ready version (non-anonymous)
\begin{document}

\maketitle
{\renewcommand{\thefootnote}{\fnsymbol{footnote}}%
\footnotetext[2]{Equal contribution; human authors listed in alphabetical order by first name.}}

\begin{abstract}
Long-context language models suffer from position bias, where information in middle positions is under-utilized. Attention Sorting addresses this by iteratively reordering documents based on attention patterns, but its multiple sort-and-generate cycles increase deployment cost. We hypothesize that position bias is the primary bottleneck and propose Debiased One-Pass Attention Sorting, which estimates a per-prompt position-bias curve from the low-attention majority of documents and uses it to correct raw attention scores (via subtraction or division) to enable single-pass sorting. Our experiments on two models refute this hypothesis in the tested setting: on LLaMA-2-7B-32K-Instruct, debiasing produces identical results to uncalibrated single-pass sorting (94.83\% containment accuracy), while on YaRN-Llama-2-7b-64k, debiasing improves accuracy by 8.67 percentage points but remains 14.84pp behind iterative sorting, closing only 37\% of the gap. These results suggest that position-bias correction is insufficient to match iterative sorting, and that repeated reordering provides additional benefits beyond bias correction.

\end{abstract}

\begin{quote}
\itshape\color{disclosurered}
\hypersetup{linkcolor=disclosurered}
\textbf{\upshape Disclosure:}\enspace
This paper was produced by FARS (Fully Automated Research System)\footnote{\url{https://analemma.ai/fars/}}, which autonomously performed the ideation, literature review, experiment design and execution, result analysis, and manuscript composition. The accompanying code is publicly available.\footnote{\url{https://gitlab.com/fars-a/calib-attnsort-onepass}}
The human authors contributed review and minor editorial revisions. They have verified the authenticity of all cited references and confirmed that all reported experimental results originate from actual code execution. Readers should be aware that the prose and presentation of this manuscript are primarily machine-generated and may not meet the standards of fully human-authored work.
\end{quote}

%%%%%%%%%INTRODUCTION%%%%%%%%%
\section{Introduction}
\label{sec:intro}

Long-context language models are increasingly deployed in retrieval-augmented generation (RAG) and multi-document question answering, where models must identify and extract relevant information from extensive contexts. However, these models exhibit systematic position bias: they tend to over-attend to information at the beginning or end of the context while under-utilizing middle positions, a phenomenon known as ``lost-in-the-middle''~\citep{Liu2023LostIT}. This bias significantly degrades performance when relevant information appears in unfavorable positions.

Attention Sorting~\citep{Peysakhovich2023AttentionSC} addresses this limitation through iterative document reordering based on attention patterns. By measuring how much attention each document receives during generation and placing high-attention documents at the end (where recency-biased models attend more effectively), the method substantially improves long-context QA accuracy. However, Attention Sorting requires $k$ iterations of sorting, each followed by answer generation, significantly increasing latency for long contexts.

We hypothesize that iterative sorting is primarily needed because raw attention scores are confounded by position bias: a relevant document far from the end may not receive sufficient attention in a single pass to be placed optimally. If we can explicitly estimate and correct this position-dependent bias, a single sorting pass should suffice to match iterative sorting accuracy. To test this hypothesis, we propose \textbf{Debiased One-Pass Attention Sorting}, which estimates a per-prompt position-bias curve from the low-attention majority of documents and uses it to correct raw attention scores (via subtraction or division), requiring only one sort pass plus one generation pass.

Our experiments on SynthWiki@28K with two models of different bias characteristics refute this hypothesis in the tested setting. On LLaMA-2-7B-32K-Instruct, debiasing produces identical results to uncalibrated $k=1$ sorting (94.83\% accuracy), with zero improvement across 600 examples. On YaRN-Llama-2-7b-64k, which exhibits severe recency bias, debiasing improves accuracy by 8.67 percentage points over uncalibrated $k=1$ but remains 14.84pp behind $k=5$ sorting, closing only 37\% of the gap.

Our contributions are threefold. First, we propose and evaluate debiased one-pass attention sorting, a method that estimates and corrects for position bias in attention scores to enable single-pass document reordering. Second, we show that this correction is unnecessary for a strong $k=1$ LLaMA-2-7B-32K-Instruct setting but beneficial for a high-bias YaRN setting, making the effect sharply model-dependent. Third, we provide evidence that position-bias correction alone does not recover iterative sorting on YaRN, suggesting that iterative sorting provides benefits beyond bias correction that future work should investigate.

%%%%%%%%%INTRODUCTION%%%%%%%%%

%%%%%%%%%RELATED WORK%%%%%%%%%
\section{Related Work}
\label{sec:related}

\subsection{Position Bias in LLMs}

Large language models exhibit systematic position bias when processing long contexts. \citet{Liu2023LostIT} identified the ``lost-in-the-middle'' phenomenon, demonstrating that LLMs perform best when relevant information appears at the beginning or end of the input, with significant performance degradation for middle positions. This U-shaped attention pattern has been attributed to intrinsic attention bias, where tokens at extreme positions receive disproportionately high attention regardless of their semantic relevance~\citep{Hsieh2024FoundIT}. \citet{Yi2025AttentionBW} characterized this as the ``attention basin'' phenomenon, showing that shallow attention layers play a critical role in determining which positions receive attention. \citet{Qiang2025InitialSaliency} further argue that initial-token saliency can contribute to U-shaped attention bias and propose directly scaling initial-token attention weights. Separately, \citet{Xiao2023EfficientSL} discovered ``attention sinks''---initial tokens that absorb excess attention mass even when semantically unimportant---which enables efficient streaming inference but also contributes to position-dependent attention allocation. Our method differs from these calibration approaches by estimating a prompt-specific document-level bias curve from the same context rather than intervening on token-level attention weights or positional encodings.

\subsection{Context Window Extension}

Extending the context window of pretrained LLMs has been an active research area. Rotary Position Embedding (RoPE)~\citep{Su2021RoFormerET} encodes absolute positions using rotation matrices while incorporating relative position information in self-attention, enabling flexible sequence lengths. However, RoPE-based models fail to generalize beyond their training sequence length. Position Interpolation~\citep{Chen2023ExtendingCW} addresses this by linearly down-scaling position indices to fit within the original context window, extending LLaMA models to 32K tokens with minimal fine-tuning. YaRN~\citep{Peng2023YaRNEC} improves upon this with a compute-efficient extension method requiring 10$\times$ fewer tokens and 2.5$\times$ fewer training steps, enabling context windows up to 128K tokens. More recently, LongRoPE~\citep{Ding2024LongRoPEEL} extends context windows to 2M tokens through progressive extension and non-uniform positional interpolation. These methods focus on enabling longer contexts but do not directly address the position bias that affects information utilization within those contexts.

\subsection{Attention-Based Document Ranking}

Several methods leverage attention patterns for document ranking in retrieval-augmented generation. Attention Sorting~\citep{Peysakhovich2023AttentionSC} iteratively reorders documents based on the attention they receive during generation, placing high-attention documents at the end where recency-biased models attend more effectively. \citet{Chen2024AttentionIL} propose in-context re-ranking (ICR), which uses attention pattern changes caused by the query for efficient zero-shot re-ranking with only two forward passes. To address inconsistencies in LLM-based ranking, \citet{Zeng2024LLMRankFusionMI} introduce LLM-RankFusion, which mitigates order and transitive inconsistencies through calibration and ranker aggregation. \citet{Tang2023FoundIT} propose permutation self-consistency, which marginalizes over different list orderings to produce order-independent rankings. More recently, \citet{Jin2025LongContextRAG} show that simple retrieval reordering can mitigate hard negatives and lost-in-the-middle effects in long-context RAG. Our work builds on Attention Sorting but investigates whether explicit position-bias correction can reduce the number of required sorting iterations.

%%%%%%%%%RELATED WORK%%%%%%%%%

%%%%%%%%%METHOD%%%%%%%%%
\section{Method}
\label{sec:method}

\subsection{Problem Setup}

We consider long-context extractive question answering, where a model receives a query $q$ and $N$ documents $\mathcal{D} = \{d_1, \ldots, d_N\}$ concatenated into a single prompt, with exactly one document containing the answer. Due to position bias~\citep{Liu2023LostIT}, LLMs tend to over-attend to documents near the beginning or end of the context while under-utilizing middle positions. For models exhibiting recency bias, placing the relevant document at the end of the context improves answer accuracy. The goal is to reorder documents so that the relevant document appears in a high-attention position (typically the end) before generating the answer.

\subsection{Attention Sorting Baseline}

Attention Sorting~\citep{Peysakhovich2023AttentionSC} addresses position bias through iterative document reordering based on attention patterns. For each document $d_i$ occupying token span $S_i$, the method computes a raw attention mass $a_i$ by aggregating attention weights from the first generated token across all layers $\ell$ and heads $h$:
\begin{equation}
a_i = \sum_{\ell=1}^{L} \sum_{h=1}^{H} \sum_{t \in S_i} A^{\ell,h}_{t},
\end{equation}
where $A^{\ell,h}_{t}$ is the attention weight from the first output token to input token $t$ at layer $\ell$ and head $h$. Documents are then sorted by ascending $a_i$, placing high-attention documents at the end. This process repeats for $k$ iterations, with an answer generated after each iteration; the final iteration's answer is used. For $k=5$, this results in 5 sort-and-generate cycles, significantly increasing latency for long contexts compared to a single generation pass.

\subsection{Debiased One-Pass Sorting}

We hypothesize that iterative sorting is primarily needed because raw attention scores $a_i$ are confounded by position bias: a relevant document far from the end may not receive sufficient attention in a single pass to be placed optimally. If we can explicitly estimate and remove this position-dependent bias, a single sorting pass should suffice.

\begin{figure}[t]
\centering
\includegraphics[width=\textwidth]{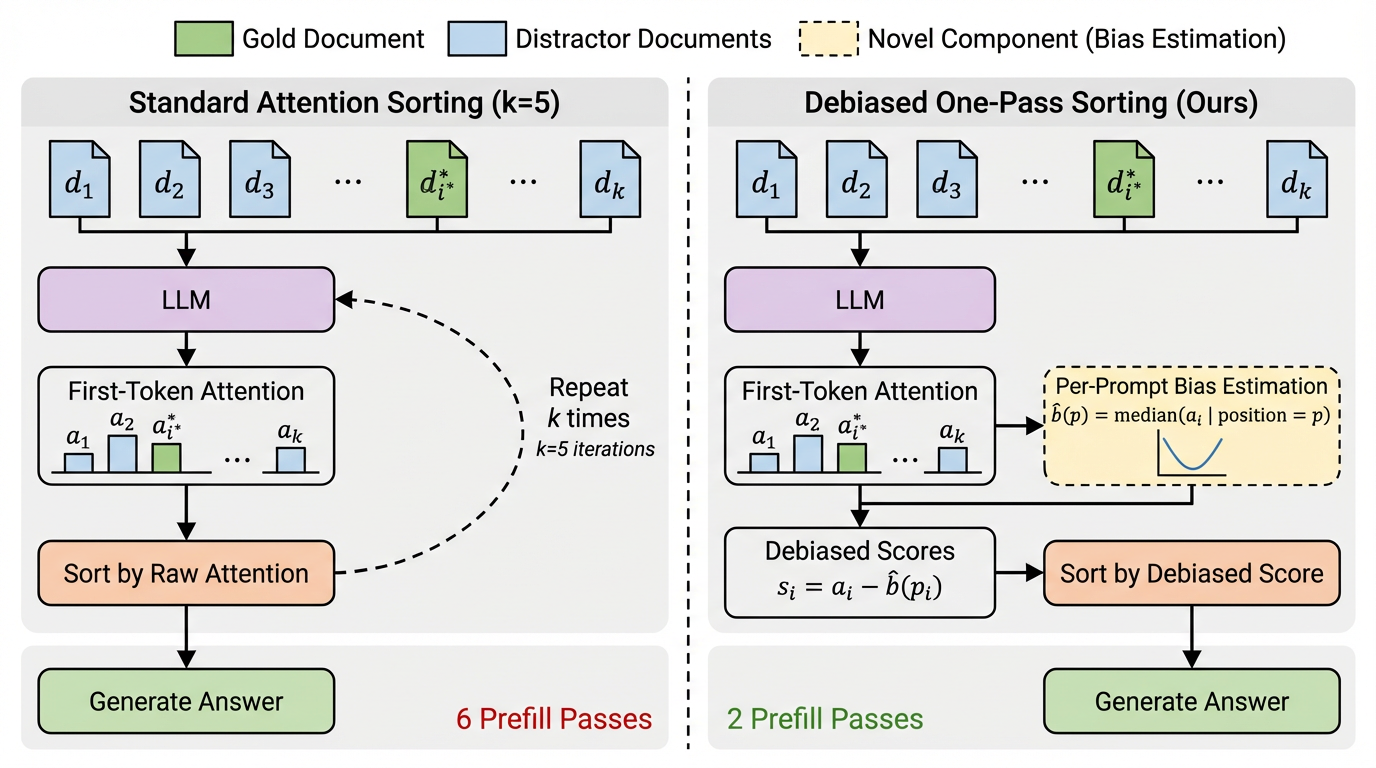}
\caption{Overview of standard iterative Attention Sorting ($k=5$, 5 sort-and-generate cycles) versus the proposed Debiased One-Pass approach ($k=1$, 1 sort + 1 generation pass). The proposed method estimates a position-bias curve from the low-attention majority of documents and uses it to correct raw attention scores for document reranking.}
\label{fig:framework}
\end{figure}

Our proposed method, illustrated in Figure~\ref{fig:framework}, operates as follows. First, we extract raw attention masses $a_i$ for each document using a single decode step, identical to the first iteration of standard Attention Sorting. Second, we estimate a position-bias curve $\hat{b}(p)$ from the attention-position pairs $(p_i, a_i)$ within the same prompt, where $p_i$ denotes document $i$'s position index. Under the assumption that most documents are distractors with near-zero relevance, we trim the top $\alpha$ fraction of documents by attention to exclude likely relevant documents, bin the remaining positions into $B$ equal-width bins, compute an aggregate statistic (median or mean) within each bin, and linearly interpolate to obtain a continuous bias curve $\hat{b}(p)$. Third, we compute debiased scores $s_i = a_i - \hat{b}(p_i)$ (additive mode) or $s_i = a_i / \max(\hat{b}(p_i), \epsilon)$ (divisive mode, with $\epsilon=10^{-10}$ in our implementation). Finally, we use the debiased scores to adjust the document ordering and generate the answer from the reordered prompt.

The choice of hyperparameters ($\alpha$, $B$, aggregation function, debiasing mode) and reordering strategy was tuned per model on the same evaluation data reported in our results, as described in Section~\ref{sec:experiments}. This approach requires only one sort pass plus one generation pass, matching $k=1$ Attention Sorting in compute while aiming to achieve $k=5$ accuracy through explicit bias correction rather than iterative refinement.

%%%%%%%%%METHOD%%%%%%%%%

%%%%%%%%%EXPERIMENTS%%%%%%%%%
\section{Experiments}
\label{sec:experiments}

\subsection{Experimental Setup}

We evaluate debiased one-pass attention sorting on the SynthWiki benchmark~\citep{Peysakhovich2023AttentionSC}, which contains synthetic long-context extractive QA instances with one gold document among many distractors. Following the original setup, we construct contexts of approximately 28K tokens by sampling distractor documents, replacing one with the gold document at a random position.

We test two models with different position bias characteristics: LLaMA-2-7B-32K-Instruct\footnote{\url{https://huggingface.co/togethercomputer/Llama-2-7B-32K-Instruct}}, an instruction-tuned variant of Llama~2~\citep{Touvron2023Llama2O} with a 32K context window and moderate position bias, and YaRN-Llama-2-7b-64k~\citep{Peng2023YaRNEC}, a context-extended model without instruction tuning that exhibits severe recency bias. We compare four methods: (1) No Sorting (vanilla generation), (2) Attention Sort $k=1$ (single-pass sorting by raw attention, then generation), (3) Debiased $k=1$ (our proposed method, same compute as $k=1$), and (4) Attention Sort $k=5$ (iterative sorting baseline). In the $k=5$ condition, the implementation generates an answer after each sorting iteration and uses the final iteration's answer, resulting in 5 sort-and-generate cycles.\footnote{This matches the original Attention Sorting implementation, which evaluates intermediate answers at each iteration.} We use greedy decoding and report normalized substring-match accuracy (the prediction is counted as correct if any acceptable gold answer appears as a substring of the lowercased model response) across 200 examples per seed with 3 random seeds (42, 123, 456).

\subsection{Main Results}

\begin{table}[t]
\centering
\caption{Results on SynthWiki@28K with LLaMA-2-7B-32K-Instruct. Under the selected minimal-swap configuration, debiased $k=1$ produces identical results to uncalibrated $k=1$. Best in \textbf{bold}.}
\label{tab:llama_results}
\adjustbox{max width=\textwidth}{
\begin{tabular}{lccccc}
\toprule
Method & Passes & Seed 42 & Seed 123 & Seed 456 & Mean $\pm$ Std \\
\midrule
No Sorting & 1 gen & 70.5\% & 76.0\% & 72.0\% & 72.83\% $\pm$ 2.84\% \\
Attn Sort $k=1$ & 1 sort + 1 gen & 94.0\% & \textbf{96.5\%} & 94.0\% & 94.83\% $\pm$ 1.44\% \\
Debiased $k=1$ (Ours) & 1 sort + 1 gen & 94.0\% & \textbf{96.5\%} & 94.0\% & 94.83\% $\pm$ 1.44\% \\
Attn Sort $k=5$ & 5$\times$(sort+gen) & \textbf{96.5\%} & 95.0\% & \textbf{95.0\%} & \textbf{95.50\%} $\pm$ 0.71\% \\
\bottomrule
\end{tabular}
}
\end{table}

\begin{table}[t]
\centering
\caption{Results on SynthWiki@28K with YaRN-Llama-2-7b-64k. Debiased $k=1$ improves +8.67pp over uncalibrated $k=1$ but remains 14.84pp behind $k=5$, closing only 37\% of the gap. Best in \textbf{bold}, second-best \underline{underlined}.}
\label{tab:yarn_results}
\adjustbox{max width=\textwidth}{
\begin{tabular}{lccccc}
\toprule
Method & Passes & Seed 42 & Seed 123 & Seed 456 & Mean $\pm$ Std \\
\midrule
No Sorting & 1 gen & 36.0\% & 33.5\% & 38.0\% & 35.83\% $\pm$ 2.25\% \\
Attn Sort $k=1$ & 1 sort + 1 gen & 47.0\% & 45.0\% & 49.5\% & 47.17\% $\pm$ 2.25\% \\
Debiased $k=1$ (Ours) & 1 sort + 1 gen & \underline{57.5\%} & \underline{53.5\%} & \underline{56.5\%} & \underline{55.83\%} $\pm$ 2.08\% \\
Attn Sort $k=5$ & 5$\times$(sort+gen) & \textbf{71.5\%} & \textbf{73.5\%} & \textbf{67.0\%} & \textbf{70.67\%} $\pm$ 3.33\% \\
\bottomrule
\end{tabular}
}
\end{table}

Table~\ref{tab:llama_results} presents results on LLaMA-2-7B-32K-Instruct. Debiased $k=1$ produces identical results to uncalibrated $k=1$ across all 600 examples. This does not prove that no debiasing intervention could affect this model, but shows that our prompt-level correction adds no value over raw attention sorting in this setting. The gap between $k=1$ (94.83\%) and $k=5$ (95.50\%) is only 0.67 percentage points, within one standard deviation.

Table~\ref{tab:yarn_results} shows results on YaRN-Llama-2-7b-64k, which exhibits severe recency bias. Here, debiased $k=1$ achieves 55.83\% accuracy, improving +8.67pp over uncalibrated $k=1$ (47.17\%). However, it remains 14.84pp behind $k=5$ (70.67\%), closing only 37\% of the gap. This shows that while debiasing provides meaningful improvement on high-bias models, it is insufficient to match iterative sorting.

The debiased method was configured per model. For LLaMA, we use minimal-swap with additive debiasing ($\alpha=0.05$, $B=20$, median); for YaRN, we use full-sort with divisive debiasing ($\alpha=0.005$, $B=40$, mean). These hyperparameters were selected on the same evaluation set reported here, which may overestimate the method's effectiveness on new data.

The effectiveness of position-bias correction is highly model-dependent. On LLaMA-2-7B-32K-Instruct, uncalibrated $k=1$ already places the gold document at mean position 164.86 out of approximately 166 documents, leaving minimal room for bias correction to improve. On YaRN, the severe recency bias means raw attention scores are significantly corrupted by position, making debiasing beneficial but still insufficient.

\subsection{Analysis: Iterative Sorting Progression}

\begin{figure}[t]
\centering
\includegraphics[width=0.9\textwidth]{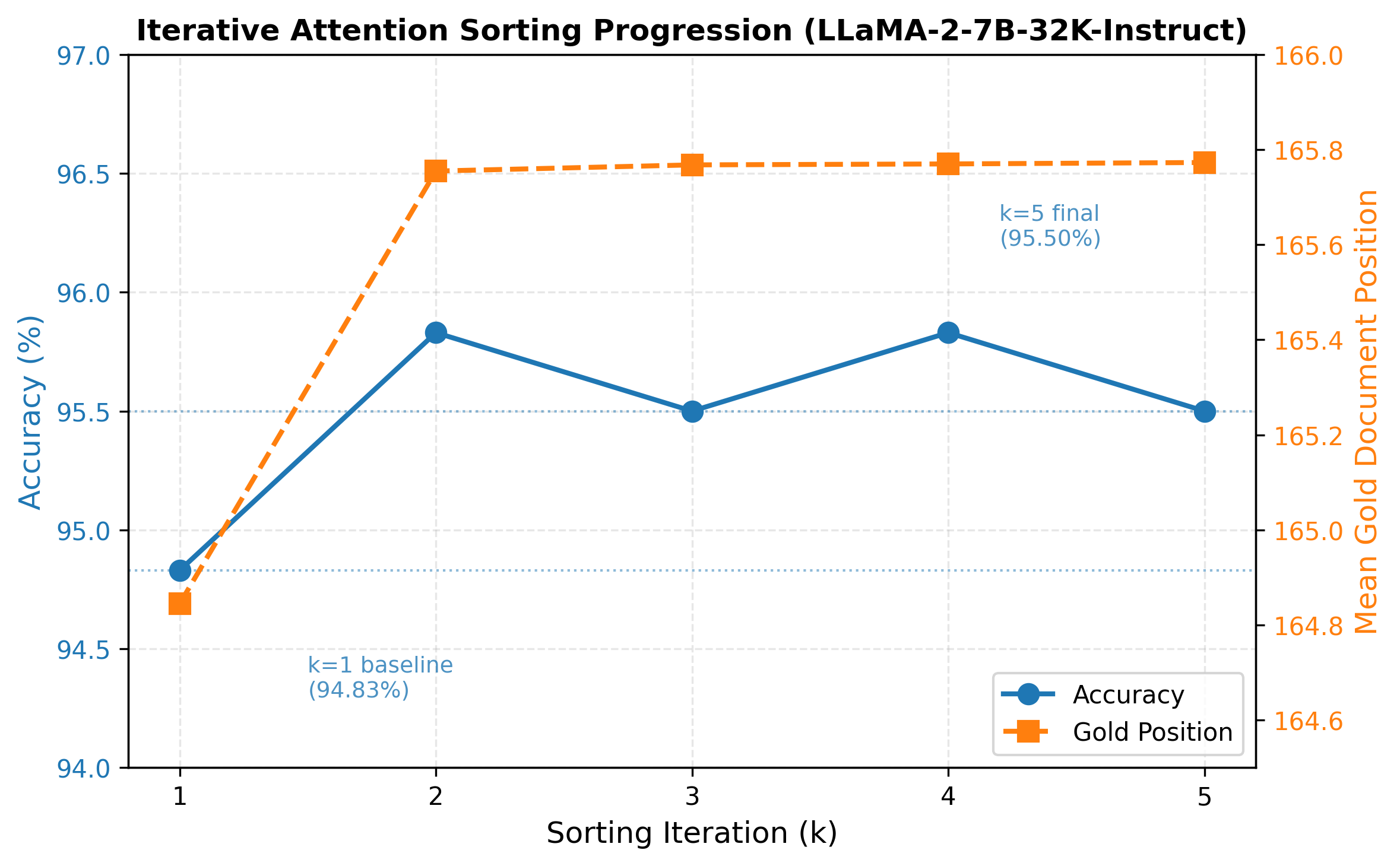}
\caption{Accuracy and mean gold document position across sorting iterations ($k=1$ to $k=5$) on LLaMA-2-7B-32K-Instruct. Most improvement occurs in the first two iterations, with accuracy increasing from 94.83\% ($k=1$) to 95.83\% ($k=2$), then plateauing. Gold document position converges to $\sim$165.77 out of $\sim$166 documents.}
\label{fig:progression}
\end{figure}

Figure~\ref{fig:progression} shows the progression of accuracy and gold document position across sorting iterations on LLaMA-2-7B-32K-Instruct. Most improvement occurs in the first two iterations: accuracy increases from 94.83\% at $k=1$ to 95.83\% at $k=2$, then plateaus with minor fluctuations through $k=5$ (final accuracy 95.50\%). The gold document position converges rapidly, reaching 164.85 after $k=1$ and stabilizing at approximately 165.77 by $k=5$.

These results suggest that iterative sorting may provide benefits beyond position-bias correction. Even after the gold document is placed near the optimal position, additional iterations continue to improve accuracy slightly, possibly from attention context refinement or noise reduction across passes. However, we do not directly measure these mechanisms, and the 0.67pp LLaMA gap is within seed variation. The stronger evidence is the 14.84pp YaRN gap after debiasing, which motivates future analysis of attention-pattern changes across sorting iterations.

%%%%%%%%%EXPERIMENTS%%%%%%%%%

%%%%%%%%%CONCLUSION%%%%%%%%%
\section{Conclusion}
\label{sec:conclusion}

We proposed debiased one-pass attention sorting, which estimates and corrects for position bias in raw attention scores to enable single-pass document reordering. Our experiments reveal that position-bias correction is insufficient to match iterative sorting: on LLaMA-2-7B-32K-Instruct, debiasing has zero effect, while on YaRN-Llama-2-7b-64k, it improves accuracy by 8.67pp but remains 14.84pp behind $k=5$ sorting. These results suggest that iterative sorting provides benefits beyond bias correction, and that future work should investigate the mechanisms underlying iterative sorting's effectiveness. Our findings indicate that debiasing should be applied selectively to high-bias models. Limitations include evaluation on only two models and one benchmark; broader evaluation across diverse models and tasks would strengthen these conclusions.

%%%%%%%%%CONCLUSION%%%%%%%%%

\bibliography{analemma}
\bibliographystyle{analemma}

%%%%%%%%%APPENDIX%%%%%%%%%
\appendix
%%%%%%%%%APPENDIX%%%%%%%%%

\end{document}